\pdfoutput=1  

\documentclass[10pt,twocolumn,letterpaper]{article}

\usepackage{cvpr}              

\usepackage{amsmath,amsfonts,bm}

\def\eqref#1{equation~\ref{#1}}

\def\1{\bm{1}}

\def\rvepsilon{{\mathbf{\epsilon}}}

\def\vs{{\bm{s}}}

\def\vv{{\bm{v}}}
\def\vw{{\bm{w}}}
\def\vx{{\bm{x}}}

\def\vz{{\bm{z}}}

\def\mA{{\bm{A}}}

\def\mV{{\bm{V}}}

\DeclareMathAlphabet{\mathsfit}{\encodingdefault}{\sfdefault}{m}{sl}
\SetMathAlphabet{\mathsfit}{bold}{\encodingdefault}{\sfdefault}{bx}{n}

\definecolor{cvprblue}{rgb}{0.21,0.49,0.74}
\usepackage[pagebackref,breaklinks,colorlinks,allcolors=cvprblue]{hyperref}

\def\paperID{1584} 
\def\confName{CVPR}
\def\confYear{2025}

\title{TCFG: Tangential Damping Classifier-free Guidance}

\author{
Mingi Kwon\thanks{Equal contribution}\\
Yonsei University\\
{\tt\small kwonmingi@yonsei.ac.kr}
\and
Shin seong Kim\footnotemark[1]\\
Yonsei University\\
{\tt\small tltydl2@yonsei.ac.kr}
\and
Jaeseok Jeong\\
Yonsei University\\
{\tt\small jete\_jeong@yonsei.ac.kr}
\and
Yi Ting Hsiao\\
University of Michigan\\
{\tt\small hsiaoyt@umich.edu}
\and
Youngjung Uh\\
Yonsei University\\
{\tt\small yj.uh@yonsei.ac.kr}
}

\usepackage{amsthm}
\newtheorem{assumption}{Assumption}
\usepackage{algorithm}
\usepackage{algorithmicx}
\usepackage{algpseudocode}
\usepackage{xcolor}
\usepackage{multirow}
\usepackage{graphicx}
\usepackage{xspace}
\usepackage{makecell} 

\usepackage{listings}
\usepackage{float} 
\usepackage{amsmath,amsfonts,bm}

\newcommand{\fref}[1]{\cref{#1}}

\newcommand{\tref}[1]{\cref{#1}}

\begin{document}
\maketitle
\begin{abstract}
Diffusion models have achieved remarkable success in text-to-image synthesis, largely attributed to the use of classifier-free guidance (CFG), which enables high-quality, condition-aligned image generation. CFG combines the conditional score (e.g., text-conditioned) with the unconditional score to control the output. However, the unconditional score is in charge of estimating the transition between manifolds of adjacent timesteps from  $x_t$  to  $x_{t-1}$, which may inadvertently interfere with the trajectory toward the specific condition. In this work, we introduce a novel approach that leverages a geometric perspective on the unconditional score to enhance CFG performance when conditional scores are available. Specifically, we propose a method that filters the singular vectors of both conditional and unconditional scores using singular value decomposition. This filtering process aligns the unconditional score with the conditional score, thereby refining the sampling trajectory to stay closer to the manifold. Our approach improves image quality with negligible additional computation. We provide deeper insights into the score function behavior in diffusion models and present a practical technique for achieving more accurate and contextually coherent image synthesis.  
\end{abstract}    

\section{Introduction}
\label{sec:intro}

Diffusion models \cite{ho2020denoising,song2020denoising} have shown remarkable progress in image generation \cite{nichol2021glide,saharia2022photorealistic,rombach2022high}. In particular, the emergence of classifier-free guidance \cite{ho2022classifier,dhariwal2021diffusion} (CFG) has attracted significant attention because it allows us to provide desired guidance by leveraging the conditional estimated score directly within the diffusion model.

\begin{figure}[!t]
\begin{center}
    \centering
    \includegraphics[width=\linewidth]{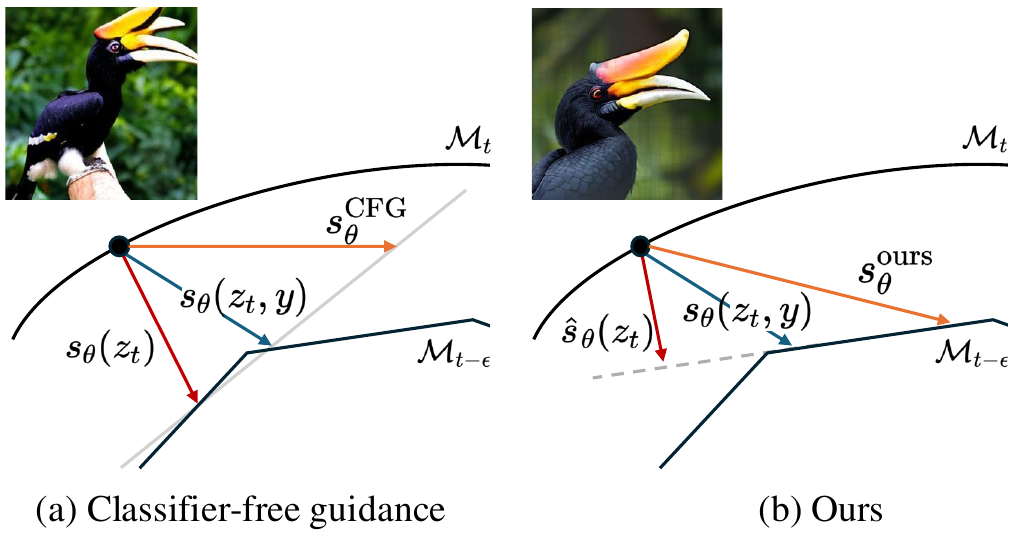}
    \caption{
    (a) Classifier-free guidance. When the unconditional score $\vs_\theta(\vz_t)$ and the conditional score $\vs_\theta(\vz_t,y)$ are misaligned, the result of CFG tends to fall off the manifold. (b) Our proposed method reduces the misalignment between the unconditional score  $\vs_\theta(\vz_t)$ and the conditional score $\vs_\theta(\vz_t,y)$, ensuring sampling aligns with the target manifold. 
    }
    \label{fig:concept_overview}
    \vspace{-2em}
\end{center}
\end{figure}

The classifier-free guidance fundamentally computes the final score by combining the unconditional and conditional estimated scores. This approach ensures a generation that aligns well with the given condition. Additionally, using an appropriate guidance scale has been shown to enhance image quality across various tasks, further driving improvements in applications like text-to-image generation. \\

Let us say the guided score $\Tilde{\vs}_\theta$ as $\Tilde{\vs}_\theta = \vs_\theta^{\text{uncond}} + \omega_{\text{scale}} (\vs_\theta^{\text{cond}} - \vs_\theta^{\text{uncond}})$.
In text-to-image models, the text condition ($\vs_\theta^{\text{cond}}$) is randomly replaced with a null condition ($\vs_\theta^{\text{uncond}}$) during training (e.g., with a probability p = 0.1), enabling the null condition to act as a general estimator for any sample. It means that $\vs_\theta^{\text{uncond}}$ is the score estimated from $z_t$ to $z_{t-1}$ for all samples in the sampling trajectory of diffusion models.

The unconditional score for any sample has certainly enabled the successful use of classifier-free guidance. However, we argue that there can be a misalignment between the unconditional and conditional estimated scores (See \cref{eq:2} in \cref{sec:background}), which hinders the approximation toward the manifold by the given condition. \fref{fig:concept_overview} (a) conceptually illustrates the potential issue that arises when the manifold of the unconditional score differs from that of the conditional score. In this paper, we show that this misalignment can be resolved with a simple algorithm, which significantly reduces the tendency of CFG to generate off-manifold samples, as illustrated in \fref{fig:concept_overview} (b).


Our approach is based on the following insights. First, the score predicted by the diffusion model
estimates the intrinsic dimension of the data manifold \citep{stanczukdiffusion}. Additionally, this intrinsic dimension
can be captured by the tangent space of the target manifold \citep{fefferman2016testing, RepresentationLearning}. Instead of directly estimating the intrinsic dimension, we focus on utilizing the \textit{tangential component} inherent in the unconditional score during classifier-free guidance. By reducing its misalignment with the conditional score, we enhance the alignment and ultimately improve the quality of the generated outputs.

Specifically, we push the score $\Tilde{\vs}_\theta$ toward the normal direction of the conditional manifold
by eliminating the values of column vectors with small singular values using the orthogonal matrix $V$ obtained through the singular value decomposition of the conditional and unconditional scores.

In this paper, we propose a novel sampling method that leverages the unconditional score within CFG. To support our approach, we first lay out the theoretical foundation in section \cref{sec:background} and \cref{sec:motivation}, discussing the manifold hypothesis and its connection to diffusion models. In \cref{sec:method}, we provide a comprehensive explanation of our proposed method. This is followed by a detailed analysis using a toy example in \cref{sec:toy}, and we demonstrate the practical applicability of our method on real-world text-to-image models in \cref{sec:experiments}.

Our experiments show a significant improvement in the MS-COCO Fréchet Inception Distance (FID) across various models that utilize classifier-free guidance, e.g., diffusion models (Stable Diffusion v1.5 \citep{Rombach_2022_CVPR} and SDXL \citep{podell2023sdxl}) and rectified flow (Stable Diffusion 3 \citep{sd3}).
Additionally, our method improves DiT \citep{peebles2023scalable} FID on ImageNet. Notably, our method helps mitigate the overexposure bias problem, leading to resulting images that better align with the underlying data distribution, as supported by improved quantitative metrics.

\section{Background}
\label{sec:background}
\paragraph{Diffusion models}

Diffusion models 
learn the score that reverses the forward noising process. This forward process from the real data distribution $p(\vx_0)$ to a latent distribution $p(\vz_1) \sim N(0, \sigma_{\text{max}}^{2} I)$ along timesteps $t \in [0, 1]$ is defined by a Gaussian kernel: $\vz_t = \vx_0 + \sigma(t) \rvepsilon$. The function \( \sigma(t) \) is a noise schedule where \( \sigma(0) = 0 \) and \( \sigma(1) = \sigma_{\text{max}} \), determining the amount of noise to be added at each timestep \( t \) to erase information from \( x \).

A generative process is represented as its reverse with a stochastic differential equation (SDE):
\begin{align}
\mathrm{d} \vz &= -\dot{\sigma}(t) \sigma(t) \nabla_{\vz_t} \log p_t(\vz_t) \, \mathrm{d} t \notag \\
&\quad - \beta(t) \sigma(t)^2 \nabla_{\vz_t} \log p_t(\vz_t) \, \mathrm{d} t + \sqrt{2 \beta(t)} \sigma(t) \, \mathrm{d} \omega_t, \notag
\end{align}
where \( \mathrm{d} \omega_t \) is a standard Wiener process.
Alternatively, it can be expressed as an ordinary differential equation:
\[
\mathrm{d} \vz = -\dot{\sigma}(t) \sigma(t) \nabla_{\vz_t} \log p_t(\vz_t) \, \mathrm{d} t.
\]
Diffusion models approximate the score function \( \nabla_{\vz_t} \log p_t(\vz_t) \) with a neural network \( \vs_\theta(\vz_t, t) \). They are trained to predict the clean data from the noisy \( \vz_t \). The trained model performs the reverse process using:
\[
\nabla_{\vz_t} \log p_t(\vz_t) \approx \frac{\vs_{\theta}(\vz_t, t) - \vz_t}{\sigma(t)^2}.
\]
\paragraph{Classifier guidance (CG) and classifier-free guidance (CFG)}
For an arbitrary class label \( y \), CG defines the class-conditional sampling distribution \( \tilde{p}_\theta(\vz_t \mid y) \) as:
\[
\tilde{p}_\theta(\vz_t \mid y) \propto p_\theta(\vz_t \mid y) \, p_\theta(y \mid \vz_t)^\gamma,
\]
where $p_\theta(y \mid \vz_t)$ is the classifier distribution and \( \gamma \) is a scaling parameter. \cite{dhariwal2021diffusion} When \( \gamma > 0 \), it is known to reduce sample diversity but enhance quality.
However, CG requires a classifier that can predict label \( y \) from the noisy \( \vz_t \). CFG proposes a method to sample from the conditional distribution by expressing the classifier distribution \( p_\theta(y \mid \vz_t) \) in terms of the conditional distribution \( p_\theta(\vz_t \mid y) \) and the unconditional distribution \( p_\theta(\vz_t) \):
\[
\tilde{p}_\theta(\vz_t \mid y) \propto p_\theta(\vz_t \mid y)^{1+\gamma} \, p_\theta(\vz_t)^{-\gamma}.
\]
As a result, the final score \( \nabla_{\vz_t} \log \tilde{p}_\theta(\vz_t \mid y) \) is approximated by:
\begin{align}
\nabla_{\vz_t} \log \tilde{p}_\theta(\vz_t \mid y) & = (1 + \gamma) \, \vs_\theta(\vz_t, y) -\gamma \, \vs_\theta(\vz_t) \notag \\
& = \vs_\theta(\vz_t) + \omega \, (\vs_\theta(\vz_t, y)-\vs_\theta(\vz_t)), \notag 
\end{align}
where $\omega=1+\gamma$. \cite{ho2022classifier}

In practice, both \( \vs_\theta(x_t, y) \) and \( \vs_\theta(\vz_t) \) are approximated by a single neural network that is jointly trained to estimate both the conditional and unconditional scores. Text-to-image models use the null condition \( y_{\text{null}}=\varnothing \) as a class label to train $\vs_\theta(\vz_t) \approx \vs_\theta(\vz_t, y_{\text{null}})$. This approach allows \( \vs_\theta(\vz_t, y) - \vs_\theta(\vz_t) \) to provide guidance similar to the gradient of an implicit classifier.  Hereafter, we will simply denote $\vs_\theta(\vz_t, y_{\text{null}})$ as $\vs_\theta(\vz_t)$.

\paragraph{Diffusion models and data manifold encoding}


The manifold hypothesis suggests that high-dimensional data lies on or near a lower-dimensional manifold, making intrinsic dimension estimation essential for data representation \citep{fefferman2016testing}. This intrinsic dimension is often encoded in the manifold’s tangent spaces, which capture underlying degrees of freedom and align local structures to reveal the global geometry \citep{RepresentationLearning, zhang2004principal}.




Building on these ideas, further studies have analyzed the approximation and generalization capabilities of diffusion models \citep{oko2023diffusion,pope2021intrinsic,pidstrigach2022score}, and have also proven that their score functions can approximate the tangent space of the data manifold \citep{stanczukdiffusion}. In particular, for a compact embedded sub-manifold $\mathcal{M} \subset \mathbb{R}^n$, it has been shown that for a sample $\vz_{t} \in \mathbb{R}^n$ sufficiently close\footnote{Every compact embedded submanifold of $\mathbb{R}^d$ has a tubular neighborhood, and for a given manifold $\mathcal{M}$, each point $\vz \in \mathbb{R}^n$ within this tubular neighborhood has a unique orthogonal projection $\pi$ onto $\mathcal{M}$ \citep{lee2018introduction}.} to the target data, the score $\nabla_{\vz_t} \log p_t(\vz_t)(\approx \vs_{\theta}(\vz_{t}))$ and orthogonal projection $\pi(\vz_{t})$ onto data manifold $\mathcal{M}_0$ satisfy a key relationship. For the projection $\mathbf{N}_{p}$ onto the normal space and $\mathbf{T}_{p}$ onto the tangent space of $\mathcal{M}_0$, the ratio of their magnitudes goes to zero as $t$ approaches 0 (i.e., gets closer to the target data). In other words, for samples $\vz_{t}$ close to the target data, the following equation holds:
\begin{align}
\frac{\| \mathbf{T}_{p} \nabla_{\mathbf{z}_t} \log p_t(\vz_{t}) \|}{\| \mathbf{N}_{p} \nabla_{\mathbf{z}_t} \log p_t(\vz_{t}) \|} &\to 0, \quad \text{as } t \to 0,
\end{align}

where $\mathbf{T}_{p}$ and $\mathbf{N}_{p}$ are the projection operators onto the tangent space $\mathcal{T}_{\pi(\vz_{t})}\mathcal{M}_0$ and the normal space $\mathcal{N}_{\pi(\vz_{t})}\mathcal{M}_0$, respectively (for a detailed proof, see Theorem 4.1, Corollary 4.2 and Appendix D in \citep{stanczukdiffusion}).

This implies that, for samples sufficiently close to the target manifold $\mathcal{M}_0$, the cosine similarity between the score function and the normal vector $\mathbf{n} = \frac{\pi(\vz_{t}) - \vz_{t}}{\|\pi(\vz_{t}) - \vz_{t}\|}$  converges to 1 (i.e., $S_{\cos}(\mathbf{n}, \nabla_{\vz_{t}} \log p_t(\vz_{t})) \xrightarrow{t \to 0} 1$).

This suggests that for \textit{a sample $\vz_{t}$ very close to the target, the score function $\nabla_{\mathbf{z}} \log p_t(\vz_{t})\approx \vs_{\theta}(\vz_t)$ becomes an element of the normal space of the target manifold} (that is, $\nabla_{\mathbf{z}} \log p_t(\vz_{t}) \in \mathcal{N}_{\pi(\vz_{t})} \mathcal{M}_0 \approx \mathcal{N}_{\pi(\vz_{0})} \mathcal{M}_0$ for sufficiently small $t$).  Leveraging this property, the estimated diffusion score can approximate the intrinsic dimension of the target data by utilizing the huge gap in the singular values of the sampling scores 
$S = \left[ \vs_\theta \left( \vz_{t}^{(1)}, t \right), \ldots, \vs_\theta \left( \vz_{t}^{(4n)}, t \right) \right]$,
where the singular vectors corresponding to the higher singular values represent the normal components of \(\mathcal{M}_0\), while those corresponding to the lower singular values represent the tangential components.\citep{stanczukdiffusion}

\section{Intuition}
\label{sec:motivation}
In this section, we assume the mathematical concept behind our method and supporting experiments. Our approach refines CFG at each step by dropping the tangential component of the unconditional score, enhancing the quality of conditional generation. This adjustment allows the conditional score to guide the generated sample more directly toward the manifold specified by the condition, improving alignment.


To support this, we provide empirical evidence suggesting that not only does the target data manifold $\mathcal{M}_{0}$ exist but there is also a manifold $\mathcal{M}_{t-\epsilon}$ at each time step $t \in (0,1)$ where $\nabla_{\mathbf{z}} \log p_t(\vz_{t}) \in \mathcal{N}_{\pi_{t-\epsilon}(\vz_t)}\mathcal{M}_{t-\epsilon}$.

\paragraph{There exists an intermediate manifold $\mathcal{M}_t$}

We hypothesize the existence of a manifold $\mathcal{M}_{(t-\epsilon)}$ that contains $\nabla_{\vz_{t}}\log p_t(\vz_t)$ as elements of its normal space, not only for samples close to the target data but also for $t \in (0,1)$. Specifically, we assume the following:

\begin{assumption}
Suppose that the support of the data distribution $P_0$ is contained in a compact embedded submanifold $\mathcal{M}_0 \subset \mathbb{R}^d$, and let $P_t$ be the distribution of latents at time $t$ diffused from $P_0$. Then, under mild assumptions\footnote{1) The distribution $P_0$ has a smooth density $p_0$ w.r.t the volume measure on the manifold.
2) The density $p_0$ is bounded away from zero on the manifold.}, $\forall t \in (0,1)$, $\exists t' \in (t-\epsilon, t+\epsilon)$ such that:
\[
\nabla_{\vz_t} \log p_t(\vz_t) \in \mathcal{N}_{\pi_{t'}(\vz_t)} \mathcal{M}_{t'},
\]
\end{assumption}
\noindent for sufficiently small $\epsilon$ and orthogonal projection $\pi_{t'}(\vz_{t})$ onto manifold $\mathcal{M}_{t'}$. This hypothesis is indirectly supported by the clear gap in singular values arranged in descending order for a sufficient number of samples. This phenomenon occurs not only when $t$ goes to 0 (near the image manifold) but also consistently for all time step $t \in (0,1)$.\\

\begin{figure}[!t]
\begin{center}
    \centering
    \includegraphics[width=\linewidth]{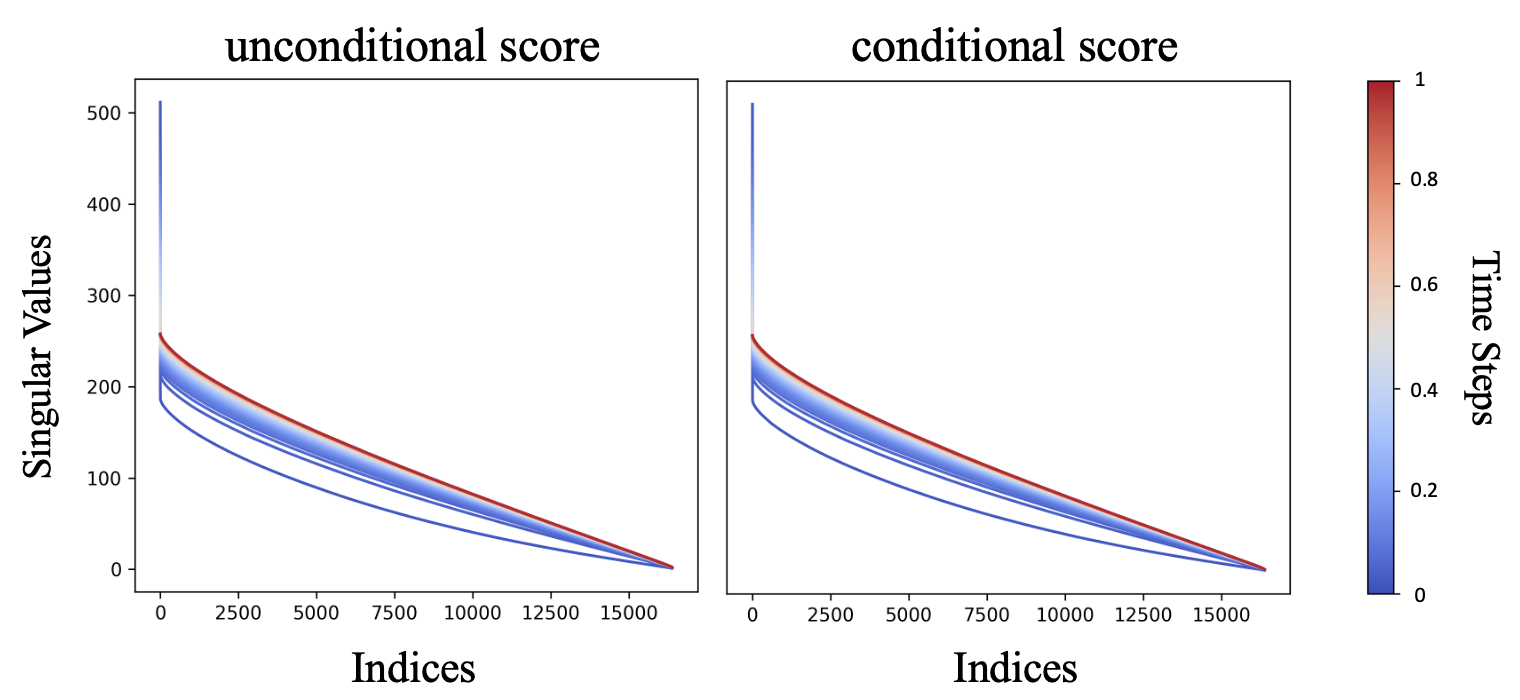}
    \caption{
    \textbf{Singular values of the score function across all timesteps.}
    We computed the singular values for all timesteps using a total of 17,000 samples from Stable Diffusion v1.5. For both the unconditional and the conditional scores, a significant drop in singular values was observed at indices close to 0 across all timesteps. This suggests the existence of an intermediate manifold.
    }
    \label{fig:behavior}
    \vspace{-2em}
\end{center}
\end{figure}


To observe the gap, we compute 17,000 score samples across all timesteps on Stable Diffusion v1.5. Let $[\sigma_1, \ldots, \sigma_{D}]$ represent the singular values from the SVD applied to $\vs_{\theta}(\vz_t)$ and $\vs_{\theta}(\vz_t,y)$, with 17,000 samples collected per timestep, arranged in descending order. The corresponding singular vectors are denoted as $[\vv_1, \ldots, \vv_{D}]^T$ for $\vs_{\theta}(\vz_t)$ and $[\hat{\vv}_1, \ldots, \hat{\vv}_{D}]^T$ for $\vs_{\theta}(\vz_t, y)$, respectively ($D \approx$\footnote{Approximately $4 \times n$ samples are sufficient to accurately estimate the intrinsic dimension of the target manifold $\mathcal{M}_0$ \cite{stanczukdiffusion}. However, our goal is to verify the existence of a manifold where $\nabla_{\vz_t} \log p_t(\vz_t)$ is an element of the normal space. Therefore, it suffices to observe the presence of a large gap in the singular value spectrum, thus $N < D$ is enough.}$17,000$). \\

As shown in \fref{fig:behavior}, both $\vs_{\theta}(\vz_t)$ and $\vs_{\theta}(\vz_t, y)$ have gaps between the highest singular values and the rest for all $t \in [0,1]$, not just for $0+\epsilon$. Interpreting from the perspective that the score function $\vs_{\theta}(\vz_t)$ becomes an element of the data manifold's normal space as $t$ approaches 0 \citep{stanczukdiffusion}. Assuming the existence of an intermediate manifold $\mathcal{M}_{t}$ for all $t \in (0,1)$, this suggests that the singular vectors associated with the largest singular values contain dominant components of $\mathcal{N}\mathcal{M}_{t}$, while vectors associated with smaller singular values correspond to component of $\mathcal{T}\mathcal{M}_{t}$.



\paragraph{Tangential misalignment between unconditional and conditional score}
We empirically justify the principle of
modifying the unconditional score by dropping the components with low singular values and retaining only the components with high singular values.


\begin{figure}[!t]
\begin{center}
    \centering
    \includegraphics[width=0.8\linewidth]{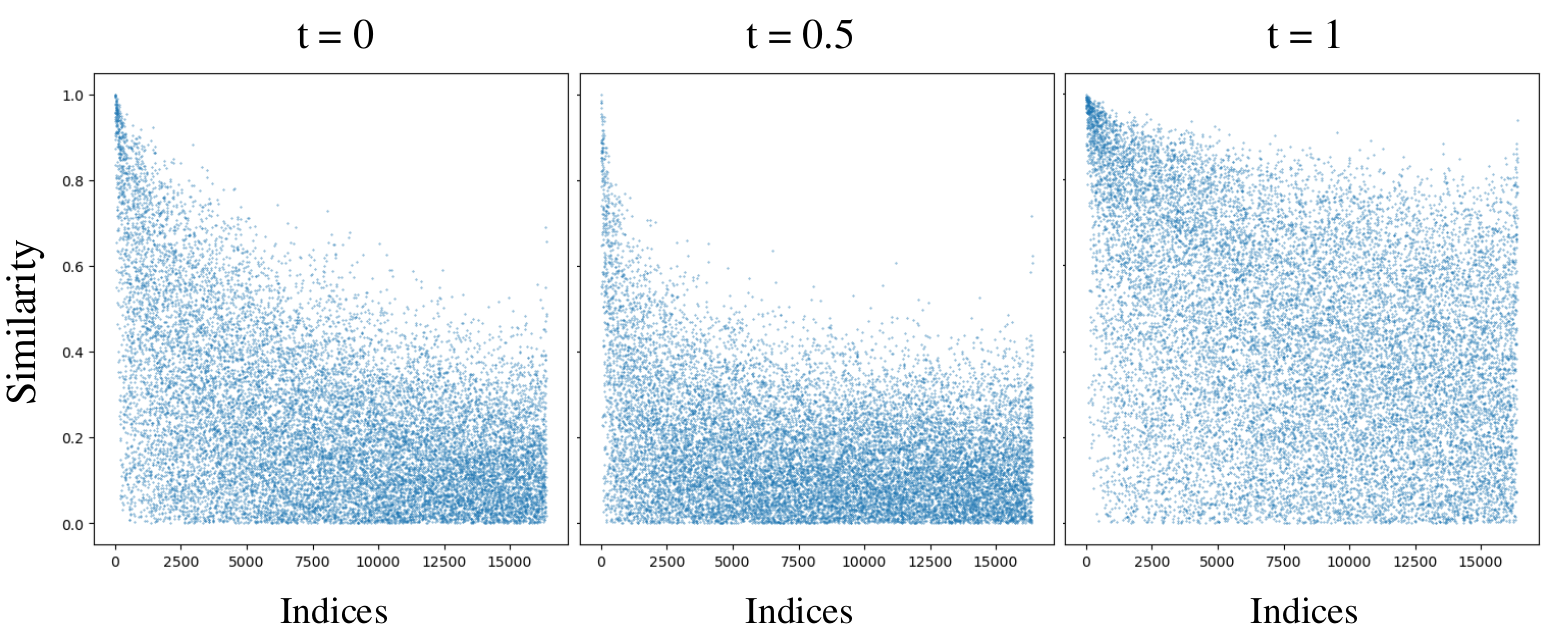}
    \caption{
    \textbf{Cosine similarity between singular vectors of unconditional and conditional scores.}
We computed the singular vectors $V$ at each timestep using a total of 17,000 samples from Stable Diffusion v1.5. We observe the similarity of significant singular vectors (i.e., those with indices close to 0) between unconditional and conditional scores are mostly high across all timesteps $T$.
    }
    \label{fig:null_cond_similarity}
    \vspace{-2em}
\end{center}
\end{figure}

\cref{fig:null_cond_similarity} shows that conditional and unconditional singular vectors $[{\vv}_1, \ldots, \vv_{D}]^T$ and $[\hat{\vv}_1, \ldots, \hat{\vv}_{D}]^T$ at corresponding indices are more similar when their singular values are high than the rest.

More specifically, the cosine similarity of the singular vectors \( \vv_1 \) and \( \hat{\vv}_1 \) associated with the highest singular value \( \sigma_1 \) from \( \vs_{\theta}(\vz_t) \) and \( \vs_{\theta}(\vz_t, y) \), respectively, is higher than the others.
\begin{equation}
\label{eq:2}
\begin{aligned}
& [S_{\cos}(\vv_{1}, \hat{\vv}_{1}) > S_{\cos}(\vv_j, \hat{\vv}_j)] \\
& \approx [S_{\cos}(\mathbf{N}_{p} \nabla_{\vz_t} \log p_t(\vz_t, y), \mathbf{N}_{p}\nabla_{\vz_t} \log p_t(\vz_t)) \\
& > S_{\cos}(\mathbf{T}_{p}\nabla_{\vz_t} \log p_t(\vz_t, y), \mathbf{T}_{p}\nabla_{\vz_t} \log p_t(\vz_t))]
\end{aligned}
\end{equation}
for $1 < j \leq D$. The cosine similarity $S_{cos}$ between two vectors $\vv_{i}$ and $\vv_{j}$ is defined as $S_{\cos}(\vv_i, \vv_j) = \frac{\vv_i \cdot \vv_j}{\|\vv_i\| \|\vv_j\|}$.\\

This indicates that the intermediate manifolds associated with \( \nabla_{\vz_t} \log p_t(\vz_t) \) and \( \nabla_{\vz_t} \log p_t(\vz_t, y) \) share similar normal components, while their tangent components are relatively less aligned.

These less-aligned components interfere with the generative process, making it harder to align with the target manifold. We modify the unconditional score \( \vs_{\theta}(\vz_{t}) \) at each timestep by removing its tangential components that are less aligned with the conditional score \( \vs_{\theta}(\vz_t, y) \). We provide detailed methods in the following section.

\begin{algorithm}[b]
\caption{Tangential damping classifier-free guidance (TCFG)}
\label{alg:ours}
\textbf{Inputs:} $\mathbf{s}_\theta(\vz_{t})$ and $\mathbf{s}_\theta(\vz_{t}|y)$: predicted unconditional and conditional scores, $t\in(0,1)$: time step, $y:$ condition, $w$: CFG scale. \\
\textbf{Output:} $\vz_0$

\begin{algorithmic}[1]
\For{$t \in (0,1)$}
    \State Get $\vs_{\theta}$ from $\vz_t$
    \State Make score matrix $\mA = [\vs_{\theta}(\vz_t), \vs_{\theta}(\vz_t, y)]$
    \State $(\sigma_i)_{i=1}^d, (\vw_i)_{i=1}^d, (\vv_i)_{i=1}^d \leftarrow \text{SVD}(\mA)$
    \State $\hat{\vs}_{\theta}(\vz_t) = \vs_{\theta}(\vz_t) \cdot \mV^T \cdot [\vv_1, \mathbf{0}]$
    \State(Dropping $\mathbf{T}\nabla_{\vz_t}{\log p_t(\vx_t)}$)
    \State $\hat{\vs}_{\theta}(\vz_t,y) = \hat{\vs}_{\theta}(\vz_t) + w(\vs_{\theta}(\vz_t, y) - \hat{\vs}_{\theta}(\vz_t))$
    \State Update $\vz_t$
\EndFor
\State \textbf{Output} $\vz_0$

\smallskip
\Statex $(\sigma_i)_{i=1}^d, (\mathbf{w}_i)_{i=1}^d, (\mathbf{v}_i)_{i=1}^d$ denote singular values, left and right singular vectors respectively.
\end{algorithmic}
\end{algorithm}

\section{Methods}
\label{sec:method}
Our main method proceeds as follows. At each step, we take the predicted unconditional score $\vs_{\theta}(\vz_t)$ and the conditional score $\vs_{\theta}(\vz_t, y)$ and concatenate them into a score matrix $\mA = [\vs_{\theta}(\vz_t), \vs_{\theta}(\vz_t, y)]$. Next, we perform SVD on \( \mA \), obtaining singular values and corresponding singular vectors that consider both components \( s(\vz_{t}) \) and \( s(\vz_{t}, y) \). This results in singular vectors \([ \vv_1, \vv_2, \ldots, \vv_D ]^T\) where \(\vv_{1}\) is the normal component of both \( s(\vz_{t}) \) and \( s(\vz_{t}, y) \). We project the unconditional score onto \(\vv_{1}\) and drop the rest.
\begin{align}
\label{eq:main}
\hat{\vs}_{\theta}(\vz_t) &= \vs_{\theta}(\vz_t) \cdot \mV^T \cdot [\vv_1, \mathbf{0}].
\end{align}

Consequently, the singular vectors associated with high singular values in the score matrix $A$ retain the well-aligned, normal components of $\vs_{\theta}(\vz_{t})$ and $\vs_{\theta}(\vz_{t},y)$, while those with lower singular values represent misaligned tangential components, which we set to zero
in \cref{eq:main}
to drop these components from the unconditional score. Next, we update the score $\hat{\vs}_{\theta}(\vz_t, y)$ with classifier-free guidance (CFG):
\begin{align}
\nabla_{\vz_t}{\log \hat{p_t}(\vz_t|y)} &= \hat{\vs}_{\theta}(\vz_t) + w(\vs_{\theta}(\vz_t, y) - \hat{\vs}_{\theta}(\vz_t)).
\end{align}
We provide a detailed algorithm in \cref{alg:ours}.

Unlike traditional CFG update methods, $\nabla_{\vz_t}{\log \hat{p_t}(\vz_t|y)}$ drops tangential component from the unconditional score at each step. It prevents accumulating misaligned components from the unconditional score $\vs_{\theta}(\vz_{t})$ using the direction of the manifold defined by the given condition $y$ over time evolution. This concept is further illustrated with a simple distribution in \cref{sec:toy}, where the toy example clarifies the benefits of our methods.

\begin{figure}[!t]
\begin{center}
    \centering
    \includegraphics[width=0.75\linewidth]{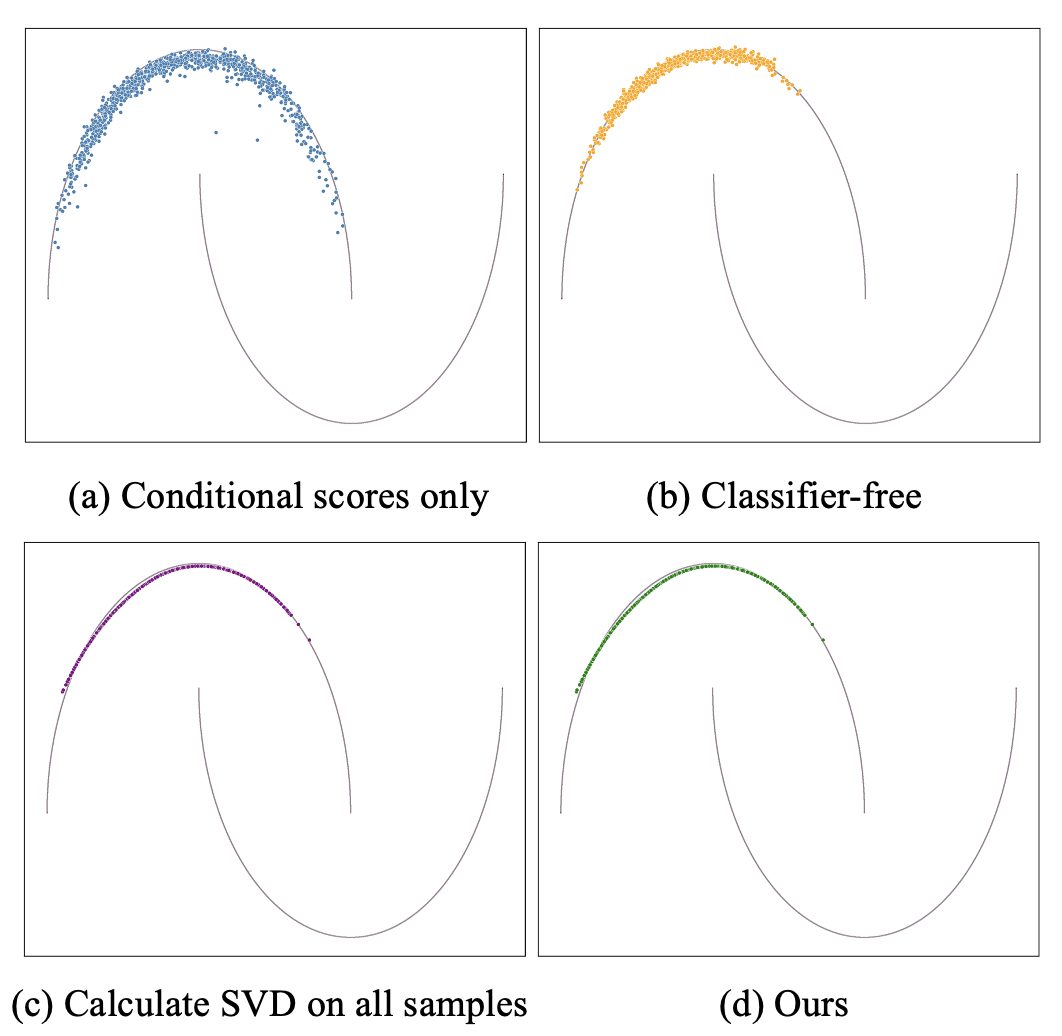}
    \caption{
    \textbf{Sampling results on different methods with diffusion model trained on \texttt{two moons} dataset}.
Our proposed methods (c, d) demonstrate a closer match to the target distribution compared to using conditional scores only or CFG. In (c), SVD is computed across all samples, while in (d), SVD is calculated separately for each pair of conditional and unconditional scores. 
    }
    \label{fig:toy_example1}
    \vspace{-2em}
\end{center}
\end{figure}

\section{Toy example}
\label{sec:toy}
We empirically verify our method on a toy problem, generating the \texttt{two moons} dataset. Experiments consist of the generated samples with different guidances including the original classifier-free guidance (CFG) and ours, and the sampling trajectories following their respective score functions.

The target data distribution $p(X_0)$ consists of samples distributed along two distinct curves (moons). We trained a conditional diffusion model using a small neural network that receives a binary label $y \in \{0,1\}$ for the two moons or $y = \varnothing$ denoting the null condition. For detailed settings, please refer to the Appendix.


\cref{fig:toy_example1} shows the generated samples
using four different guiding strategies. (a) uses only the conditional score $\vs_{\theta}(\vz_t, y)$. (b) uses the CFG score. (c) and (d) employ our guidance score at each step with multiple samples and one sample, respectively, to compute singular value decomposition (SVD) of the unconditional score $\vs^{i}_{\theta}(\vz_t)$ and conditional score $\vs^{i}_{\theta}(\vz_t, y)$.

\begin{figure}[!t]
\begin{center}
    \centering
    \includegraphics[width=\linewidth]{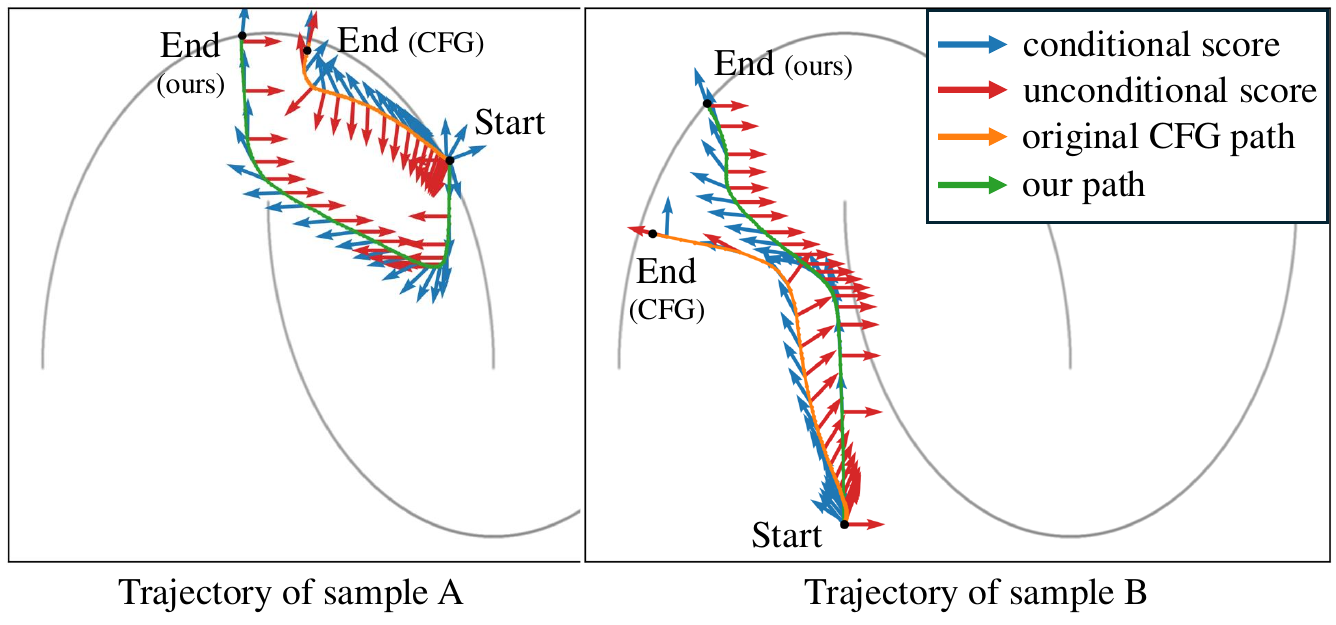}
    \caption{
    \textbf{Visualization of the sampling trajectory.} In CFG (orange path), the unconditional scores (red arrows) include components that point towards directions other than the target distribution, making the final destination deviate from the target distribution. Whereas, our method (green path) removes the inconsistent tangent components in unconditional scores and eventually reaches the target distribution.
    }
    \label{fig:toy_trajectory}
    \vspace{-2em}
\end{center}
\end{figure}

According to the result, generated samples using our strategies lie closer to the target compared to those generated using only the conditional score or CFG.
CFG, while potentially bringing samples closer to the target than merely using conditional scores, may face challenges due to the misalignment of tangent components between unconditional scorse and conditional scores.

In contrast, our guidance score can reduce the tangent component of the unconditional score at each step. This helps samples converge more effectively towards the target data, which suggests that the tangent components of the unconditional score might hinder alignment with the target data manifold under the given condition, and our method helps in mitigating this misalignment.


We further validate this hypothesis by examining the trajectories of generated samples. \fref{fig:toy_trajectory} visualizes the trajectories induced by our score \( \nabla_{\vz_t} \log \hat{p_t}(\vz_t | y) \) compared to the original CFG score \( \nabla_{\vz_t} \log \tilde{p}_{\theta}(\vz_t | y) \). As shown, in the orange CFG trajectory, the direction of unconditional scores changes frequently. This results in difficulties for the blue conditional score to maintain an orthogonal direction relative to the target manifold near the target distribution. In contrast, our method consistently adjusts the score to predict in a direction closer to orthogonal with respect to the target manifold, particularly as the samples converge toward the target data. Our method removes the tangential component of the unconditional score with respect to the manifold of the conditional score. This results in a direction that leans either to the right or to the left.

Additionally, the similar results between (c) and (d) in \fref{fig:toy_example1} suggest that computing SVD for only a single sample is sufficient to yield nearly the same result.

\section{Experiments}
\label{sec:experiments}
In this section, we demonstrate that our method is applicable to high-dimensional diffusion models. We employ representative diffusion models such as Stable Diffusion v1.5 \cite{rombach2022high} and SDXL \cite{podell2023sdxl}, and showed that it functions identically on SD v3 \cite{esser2024scaling}, which is based on Rectified Flow. Additionally, we conducted experiments on DiT \cite{peebles2023scalable}, which is trained on ImageNet \cite{deng2009imagenet}.

\paragraph{Experimental details}
For the text-to-image models, we used zero-shot FID \cite{heusel2017gans} and CLIPScore \cite{radford2021learning} on the MS-COCO 2014 validation set \cite{lin2014microsoft} consisting of 30,000 images under the commonly used text-to-image evaluation protocols. \cite{rombach2022high,podell2023sdxl,esser2024scaling} For DiT, we evaluated using 50,000 images under the same settings as ADM \cite{dhariwal2021diffusion}. All models used the official pretrained weights, and sampling was performed using the same latent codes. We used the best CFG scales as the default value of each repository. 
Our method does not increase the inference time of all baselines.

\begin{table}[!t]
\centering
\resizebox{0.6\linewidth}{!}
{%
\begin{tabular}{ll|cc}
                         &          & FID ↓  & CLIPScore ↑  \\ \hline
\multirow{2}{*}{SD v1.5} & original & 13.26 & 0.31 \\
                         & + ours   & \textbf{13.12} & 0.31 \\ \hline
\multirow{2}{*}{SDXL}    & original & 13.36 & 0.32 \\
                         & + ours   & \textbf{12.65} & 0.32 \\ \hline
\multirow{2}{*}{SD v3}   & original & 16.66 & 0.32     \\
                         & + ours   & \textbf{13.74} & 0.32 
\end{tabular}%
}
\caption{Zero-shot FID and CLIPScore measured on MSCOCO 30k. Our method consistently improves FID across all models—Stable Diffusion v1.5, SDXL, and SD v3—while maintaining a nearly identical CLIPScore.}
\vspace{-1em} 
\label{tab:t2i}
\end{table}

\begin{table}[!t]
\centering
\resizebox{\linewidth}{!}{%
\begin{tabular}{l|ccccc}
         & FID ↓  & sFID ↓ & Precision ↑ & Recall ↑ & IS ↑    \\ \hline
DiT      & 32.67 & 17.92 & \textbf{0.90}      & 0.13   & \textbf{271.1} \\
DiT+ours & \textbf{29.5}  & \textbf{13.27} & \textbf{0.90}      & \textbf{0.19}   & 270.0
\end{tabular}%
}

\caption{Evaluation metrics measured on ImageNet 50k using DiT. Our method achieves better performance in FID, sFID, Precision, and Recall while showing a slight decrease in Inception Score.}
\vspace{-2em} 
\label{tab:dit}
\end{table}

\paragraph{Quantitative evaluation}

\begin{figure}[t]
    \centering
    \includegraphics[width=0.7\linewidth]{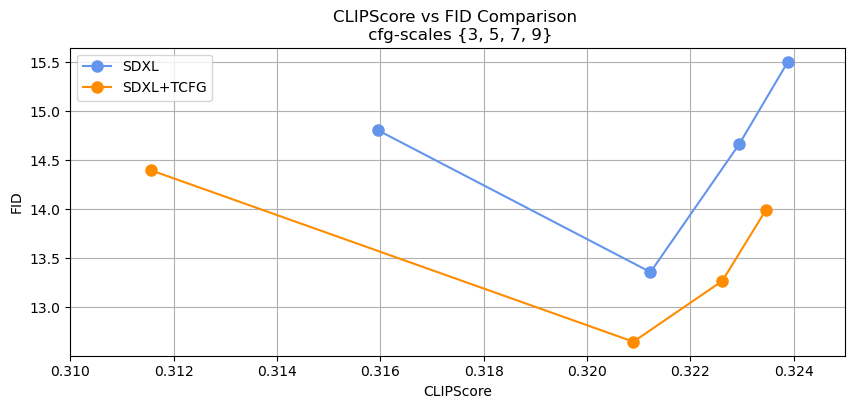}
    \vspace{-1.0em}
    \caption{FID-CLIP curves on SDXL with 50 sampling steps.
    }
    \vspace{-2em} 
    \label{fig:fid_clip}
\end{figure}

\tref{tab:t2i} presents the FID and CLIP Scores for SD1.5, SDXL, and SD3. Our method achieved better FID scores while maintaining the same CLIP Scores across all three models. Notably, the decrease in FID is larger for SDXL compared to SD1.5, and even larger for SD3 compared to SDXL. We speculate that this is because SD3, known as a better model publicly, has a relatively clearer manifold. Furthermore, the results on SD3 demonstrate that our method is applicable not only to diffusion models but also to all CFG-based score functions, including those based on Rectified Flow. \fref{fig:fid_clip} also shows FID-CLIP curves on SDXL, demonstrating that FID improves even as the CFG scale changes.

\tref{tab:dit} shows the results on the DiT model. Except for a slight decrease in Inception Score, our method exhibits relatively superior performance in FID, sFID, and Recall. This indicates that our method can be equally applied to both text-to-image generation and class-conditioned generation.

\begin{figure*}[!t]
\begin{center}
    \centering
    \includegraphics[width=\textwidth]{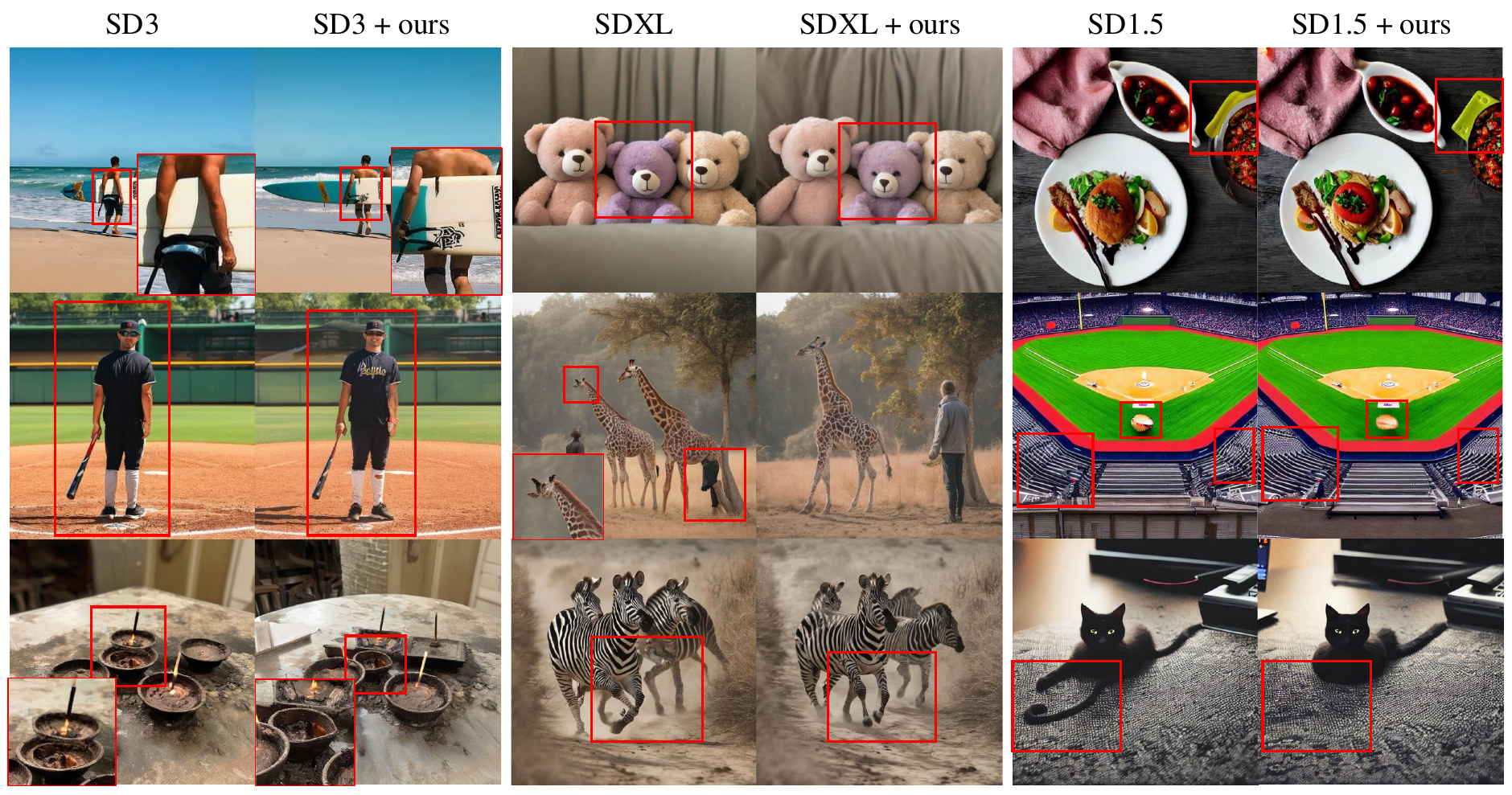}
    \caption{
    \textbf{Qualitative evaluation of text-to-image models.}
    Our method prevents overexposure, enhancing the shapes and details of objects.
    }
    \label{fig:sd_compare}
    \vspace{-3em}
\end{center}
\end{figure*}

\paragraph{Qualitative evaluation}
Our method drops the tangential component from the unconditional score while retaining the normal component. This reduces misalignment with the conditional score, thereby improving image quality as shown in \fref{fig:sd_compare}. Specifically, the changes introduced by our approach transform ``strange" objects or scenes into more ``plausible" images. This indirectly demonstrates that the misalignment of the unconditional score in the conventional CFG was causing the ``strange" aspects in the final outputs.

For example, our method converts physically impossible or unusual combinations of objects (SD3), uncommon appearances or characteristics (SDXL), and ambiguous shapes or forms (SD1.5) into ``normal" results.

\begin{figure}[!t]
\begin{center}
    \centering
    \includegraphics[width=\linewidth]{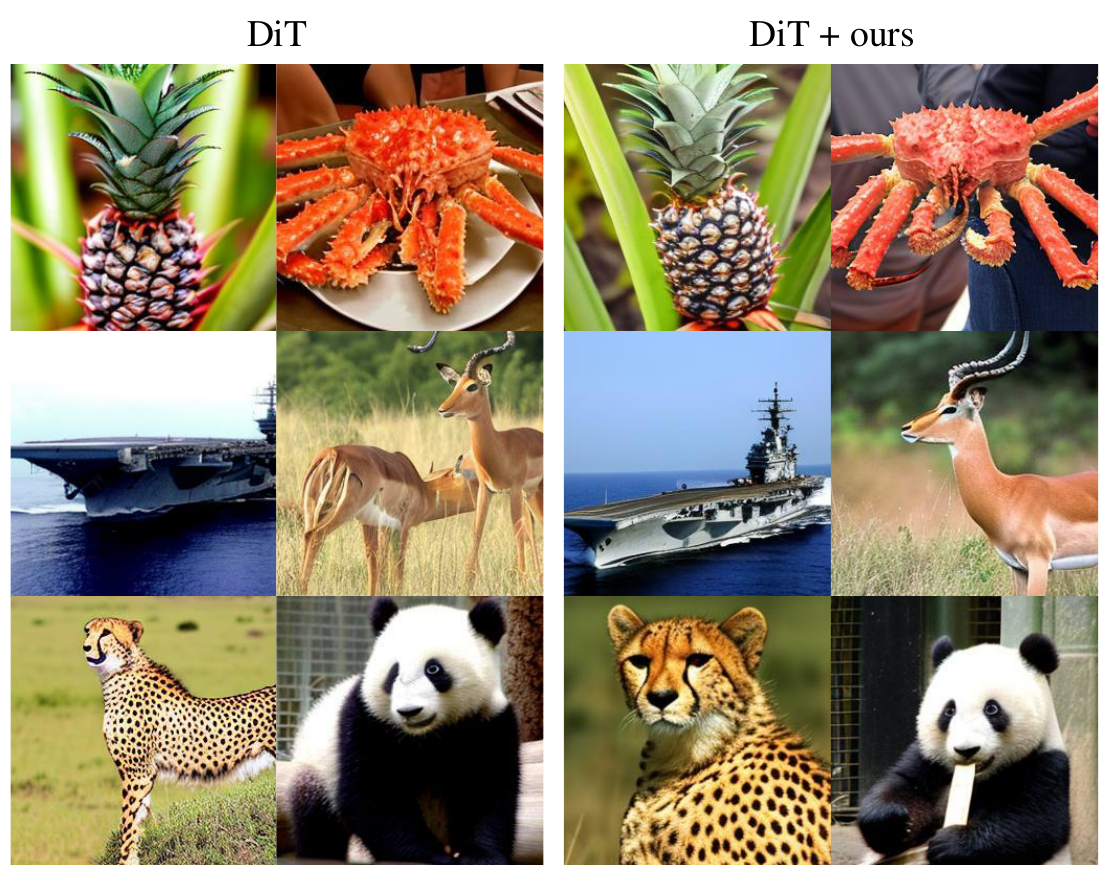}
    \caption{
    \textbf{Qualitative evaluation of DiT}
    Our method mitigates overexposure and enhances object shapes and details in DiT models trained on ImageNet.
    }
    \label{fig:dit_compare}
    \vspace{-3em}
\end{center}
\end{figure}

\fref{fig:dit_compare} presents the results obtained from DiT. We observed that our method causes relatively more changes in the images generated by DiT. We speculate that this is because DiT is trained on ImageNet dataset with class labels. The results qualitatively show that when using our method, DiT generates images that are more detailed, have better structure, and appear more natural.

Notably, in both text-to-image and class-conditioned image generation, we observed a reduction in the overexposure bias problem. We attribute this improvement to the mitigation of misalignment between the unconditional score and the conditional score.

\begin{figure*}[!t]
\begin{center}
    \centering
    \includegraphics[width=\textwidth]{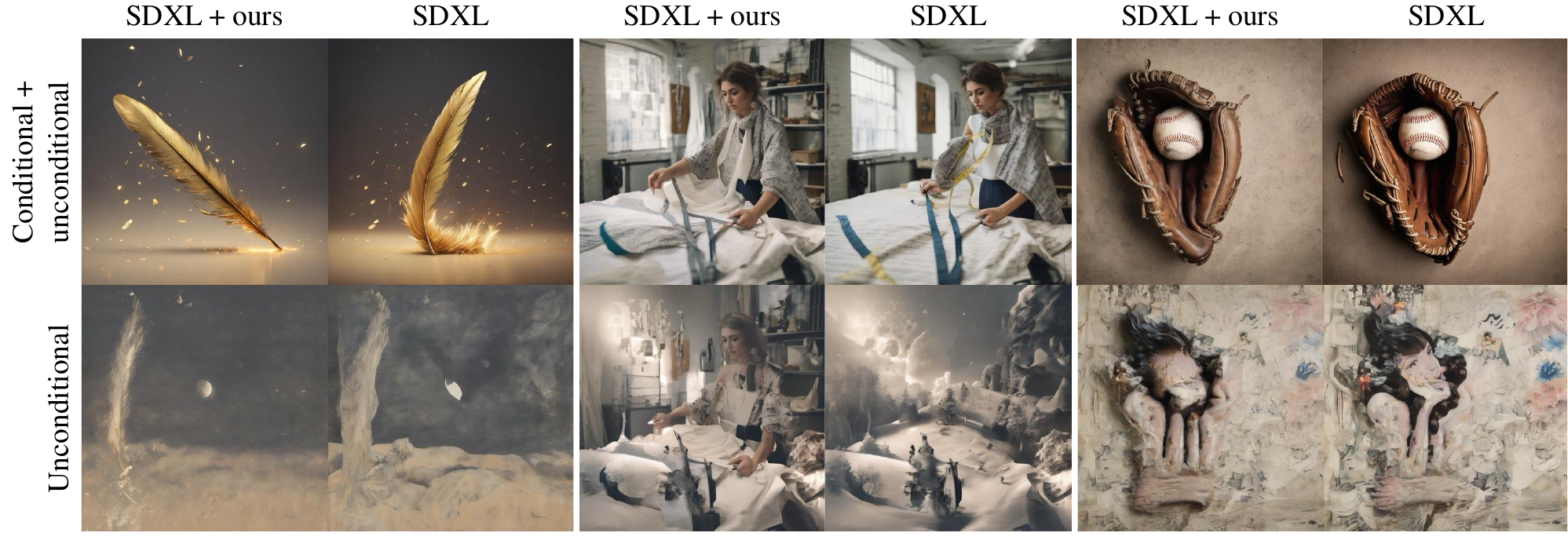}
    \caption{
\textbf{TCFG reduces misalignments between unconditional and conditional generation.} Starting from the same random noise \( \vz_1 \), when SDXL samples images with only the unconditional score, it
produces random images such as trees, snowy mountain landscapes, and women. In contrast, our modified unconditional score, projected on dominant (conditional), generates images that somewhat match the desired text prompts. This is because our method reduces misalignment with the conditional score by dropping the tangential components of the unconditional score. Once the misalignment decreases, the quality of the final images (unconditional + conditional score) improves: The base of the feather has a more natural structure, the human arm appears more natural, and the extra string on the left side of the baseball glove is removed.
    }
    \label{fig:null_visualize}
    \vspace{-3em}
\end{center}
\end{figure*}

\paragraph{What happened to the unconditional score?}

In this paragraph, we qualitatively demonstrate that the misalignment with the conditional score is reduced when we drop the tangential component from the unconditional score and retain the normal component. We compared the results sampled using CFG with those sampled using the null condition (i.e., unconditional) when generating images from the same random noise (i.e., latent variables) in SDXL. In our method, we used the text condition to compute \(\hat{\vs}\) but used only \(\hat{\vs}\) for denoising; that is, we set \(\omega = 0\). Although this approach does not perfectly explain our method, we can indirectly infer its role by observing how the modified null condition changes.

\fref{fig:null_visualize} shows that images sampled using the original null condition generate different objects such as trees, snowy mountain landscapes, and women. In contrast, images generated using our modified null condition \(\hat{\vs}\) show that the tree part takes the form of a feather, the snowy mountain landscape changes into a woman, and the woman transforms into a shape resembling a glove. These changes align with the objects we aim to generate: a feather, a woman, and a baseball glove. We observe that these changes due to the null condition help eliminate unwanted structures or artifacts in the generated images. In other words, we demonstrate that the misalignment of the null condition is reduced, and we claim that this improvement aids in image generation.

\section{Related work}

\paragraph{Calssifier-free guidance}
Experimental methods to enhance the performance of Classifier-Free Guidance (CFG) have been studied. SAG \cite{hong2023improving} proposed a method to improve CFG by using intermediate self-attention maps. PAG \cite{ahn2024self} suggested computing CFG by transforming self-attention maps into identity matrices. ICG \cite{sadat2024no} enhanced CFG by utilizing random text embeddings. Recently, CFG++ \cite{chung2024cfg++} demonstrated better performance by modifying the CFG computation method. Our proposed approach modifies the unconditional score based on the conditional score and can be used alongside these existing works; please refer to \tref{tab:comparison} and the appendix for more results.

\begin{table}[t]
\resizebox{\linewidth}{!}{%
\begin{tabular}{l|cc|cc|cc}
           & SAG           & SAG+TCFG       & PAG   & PAG+TCFG       & CFGPP & CFGPP+TCFG     \\ \hline
FID        & 13.53         & \textbf{11.48} & 14.45 & \textbf{11.87} & 13.97 & \textbf{13.44} \\
CLIP Score & \textbf{0.31} & 0.30           & 0.31  & 0.31           & 0.32  & 0.32          
\end{tabular}%
}
\vspace{-1em}
\caption{Quantitative comparison with existing baselines. The evaluation was conducted on 30k images from the MS-COCO dataset using the official code; SD v1.4 for SAG, SD v1.5 for PAG and SDXL for CFG++.
}
\vspace{-2em}
\label{tab:comparison}
\end{table}

\paragraph{Manifold hypothesis and diffusion}
There are also several studies that have utilized the manifold hypothesis properties of the score function estimated by diffusion models to address various inherent challenges associated with diffusion processes. One approach introduces the manifold memorization hypothesis to understand model memorization through the relationship between data and model manifold dimensionalities \citep{ross2024geometric}. Another extends memorization theory to diffusion models \citep{achilli2024losing}, showing that high-variance subspaces are selectively lost due to memorization effects. Separately, different researchers proposed an approach for detecting synthetic images generated by diffusion models, achieving high accuracy across diverse datasets \citep{lorenz2023detecting}.



\begin{figure}[!t]
\begin{center}
    \centering
    \includegraphics[width=0.7\linewidth]{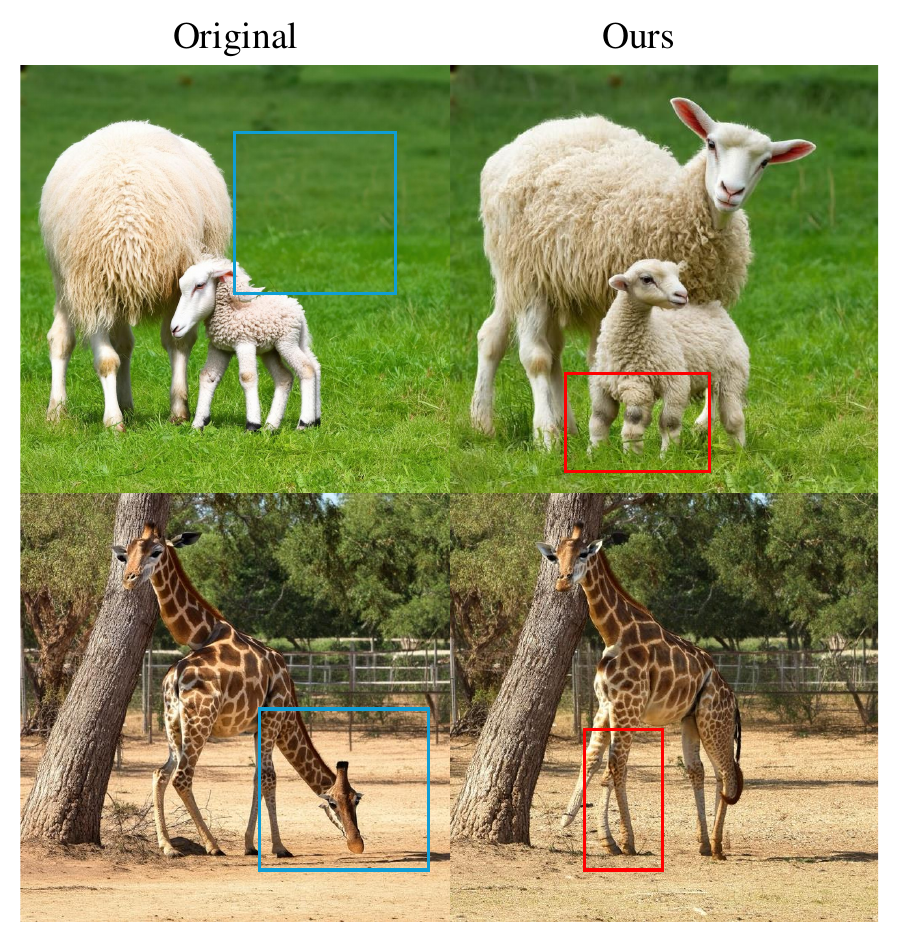}
    \caption{\textbf{Limitations}
    Our method occasionally struggles to fix severely wrong regions in the baseline samples.
    }
    \label{fig:limitation}
    \vspace{-3em}
\end{center}
\end{figure}

\section{Discussion and conclusion}

Our work experimentally analyzes the issues arising in the standard CFG method, where the tangential component of the unconditional score does not align well with that of the conditional score. By using SVD to drop the tangential component in the unconditional score, we effectively improve text-to-image generation quality. Additionally, our CFG method is easily applicable, has low computational cost, and enhances image quality. We leverage the ability of the diffusion model’s score function to encode the intrinsic dimension of the target data, demonstrating the misalignment between the conditional and unconditional scores to improve sampling quality. This is the first attempt to utilize this misalignment to enhance sampling.\\

Despite these advantages, several unresolved issues remain. First, it is uncertain whether the misalignment of tangential component and the alignment of normal component between the predicted unconditional score $\vs_\theta(\vz_t)$ and the conditional score $\vs_\theta(\vz_t, y)$ in the CFG setting, at a given timestep $t$, would similarly apply to the features derived from a separately trained classifier and the null condition score in the classifier guidance setting.
Second, while our task leverages the capability of diffusion models to estimate intrinsic dimensions for enhancing conditional sampling methods, we present only experimental observations regarding the existence of an intermediate manifold for $t \in [0, 1]$, without theoretical proof. Further exploration and rigorous analysis of these aspects are left as future work. 
Third, additional investigation is needed to adapt our approach effectively in the context of diffusion distillation using CFG scale as an input \cite{flux2024,hsiao2024plug}, which we also identify as a promising direction for future research.

Finally, although our method successfully demonstrated on-manifold image generation, we observed that when the original image exhibits significant abnormalities, substantial changes may occasionally cause the structure to break down. \fref{fig:limitation} illustrates such examples. Nevertheless, it is evident that our method transforms “strange” images into more “normal” ones.




{
    \small
    \bibliographystyle{ieeenat_fullname}
    \bibliography{main}
}

\clearpage
\setcounter{page}{1}
\maketitlesupplementary

\section{Computational Efficiency of SVD in Our Method}
\label{sec:time}
Our method requires performing SVD with only two components: the unconditional score and the conditional score. As a result, the computational time required for this operation is negligible. \tref{tab:time_diff} illustrates the additional time introduced by the SVD calculation.

The computational cost varies depending on the image resolution, as higher resolutions require larger dimensional SVD computations. For instance, the time required for SVD in SDv3 with a 1024 resolution is greater than that for SDv1.5 with a 256 resolution. However, even in the case of SDv3, the time taken remains under 0.1 seconds per image, accounting for less than a 0.01
 
 For memory usage, even with SD v3 (the largest latent dimensions), the additional memory was only 18.48 MB. In Figures 2, 3, and 4 of Section Intuition, we highlight that SVD requires only two tensors. Our design choice (full\_matrices=False during SVD) further optimizes memory, resulting in memory complexity:
$\text{Memory}_{\text{reduced}} \approx O(m + n)$.
Since  $n = 2$, memory usage scales linearly with the latent dimension  $m$.

\section{Toy Example Experiment Setup}
In the toy example experiment, we utilized the \texttt{two moons} dataset from scikit-learn. The two moons were conditioned on labels 0 and 1, while label 2 was used for the unconditional setting. The setup followed the standard DDPM configuration with 100 timesteps for training. The noise schedule employed a linear beta schedule with  $\beta_{\text{min}}$ = 0.0001  and  $\beta_{\text{max}}$ = 0.02 .
The network consisted of two linear layers, trained using the Adam optimizer with a learning rate of 0.001 for 5,000 iterations.

\section{Verifying Cosine Similarity Across Singular Vectors}

\begin{figure}[!t]
\begin{center}
    \centering
    \includegraphics[width=0.8\linewidth]{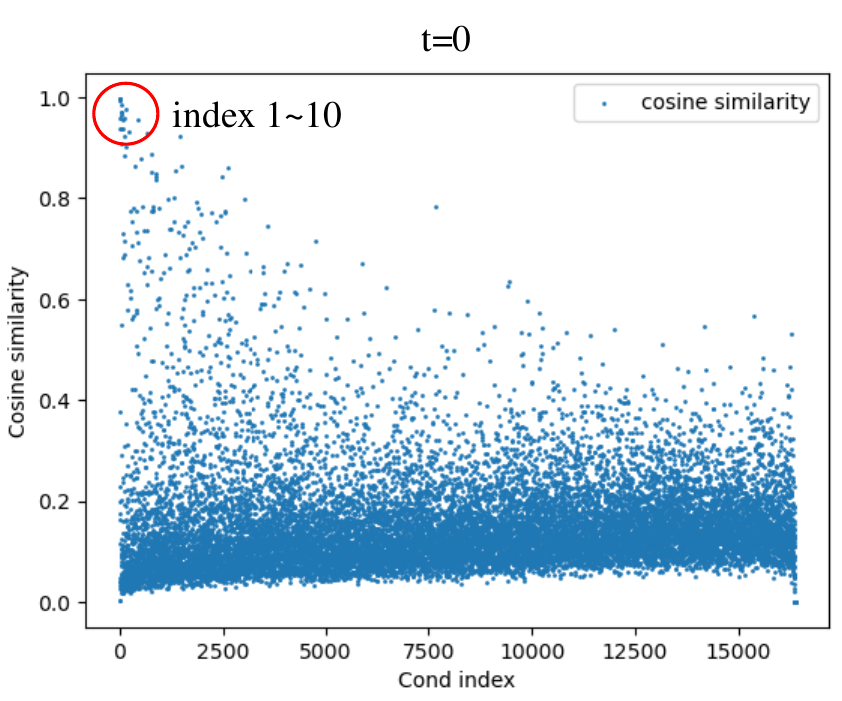}
    \caption{Cosine similarity between singular vectors of unconditional and conditional scores. We measured the cosine similarity of all 17,000 singular vectors based on the text conditional score order, ensuring that each singular vector was used only once by selecting and plotting the highest similarity value for each singular vector without duplication.}
    \label{fig:appendix_similarity}
\end{center}
\end{figure}

\fref{fig:null_cond_similarity} demonstrates that the cosine similarity between the singular vectors of the unconditional and conditional scores is significantly high for indices close to 0. However, the order of indices may differ between the unconditional and conditional scores. To ensure that the results in Figure 3 are not influenced by differing index orders, we conducted the experiment shown in \fref{fig:appendix_similarity}.

In this experiment, we measured the cosine similarity of all 17,000 singular vectors based on the text conditional score, ensuring that each singular vector was used only once by selecting and plotting the highest similarity value for each singular vector without duplication. The results consistently show that high similarity is observed only for lower indices, corroborating the findings of the original experiment. This confirms that the observed pattern is not due to the order of indices but rather reflects the fact that singular vectors corresponding to high singular values are indeed similar.

\begin{table}[t]
\resizebox{\linewidth}{!}{%
\begin{tabular}{c|c|ccc}
               & NFE                 & \multicolumn{1}{c}{\makecell{Execution \\ Time (s)}} & \multicolumn{1}{c}{\makecell{Time \\ Difference (s)}} & \multicolumn{1}{c}{\makecell{Percentage \\ Difference (\%)}} \\ \hline
SD v1.5        & \multirow{2}{*}{50} & 2.556                                              & -                                                  & -                                                           \\
SD v1.5 + ours &                     & 2.577                                              & 0.021                                              &  + 0.008                                                             \\ \hline
SDXL           & \multirow{2}{*}{50} & 13.176                                             & -                                                  & -                                                            \\
SDXL + ours    &                     & 13.221                                             & 0.045                                              &  + 0.003                                                           \\ \hline
SD v3          & \multirow{2}{*}{40} & 19.473                                             & -                                                  & -                                                            \\
SD v3 + ours   &                     & 19.558                                             & 0.085                                              &    + 0.004                                                       
\end{tabular}%
}
\caption{Comparison of execution times for standard diffusion models and our method across different resolutions and models. The additional time introduced by our method is negligible, with percentage differences remaining below 0.01\% in all cases.}
\label{tab:time_diff}
\end{table}

\section{Compatibility of Our Method with Other Techniques}
\begin{figure*}[!t]
\begin{center}
    \centering
    \includegraphics[width=\textwidth]{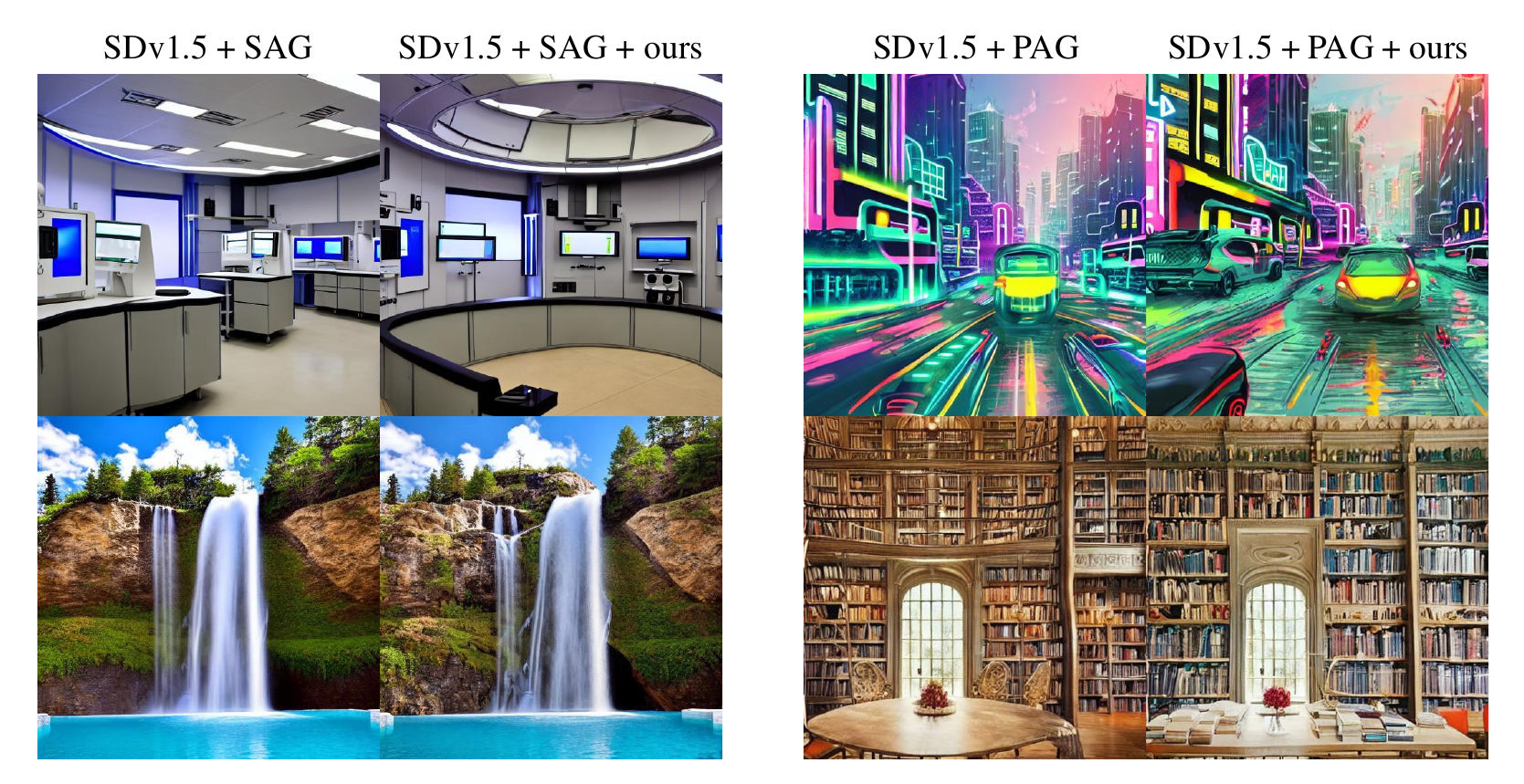}
    \caption{We observed that incorporating our method with SAG and PAG approaches improved the image structure, details, and overall color quality.}
    \label{fig:sagpag}
\end{center}
\end{figure*}

Our method modifies unconditional scores using text conditions, making it compatible with other approaches. For instance, in SAG \cite{hong2023improving}, the unconditional score is derived by blurring the attention map. We applied our projection method to the unconditional score used in SAG, and \fref{fig:sagpag} demonstrates improved results when combined with our method.

In PAG \cite{ahn2024self}, an additional score is used alongside CFG, where the self-attention map is set to identity. We observed that projecting the perturbed-attention guidance score in PAG did not yield significant improvements, likely because this score differs fundamentally from the CFG unconditional score. Instead, we projected the unconditional score used in PAG’s CFG computation using TCFG, resulting in enhanced image details and structure. Please refer to \fref{fig:sagpag}.

\begin{figure*}[!t]
\begin{center}
    \centering
    \includegraphics[width=\textwidth]{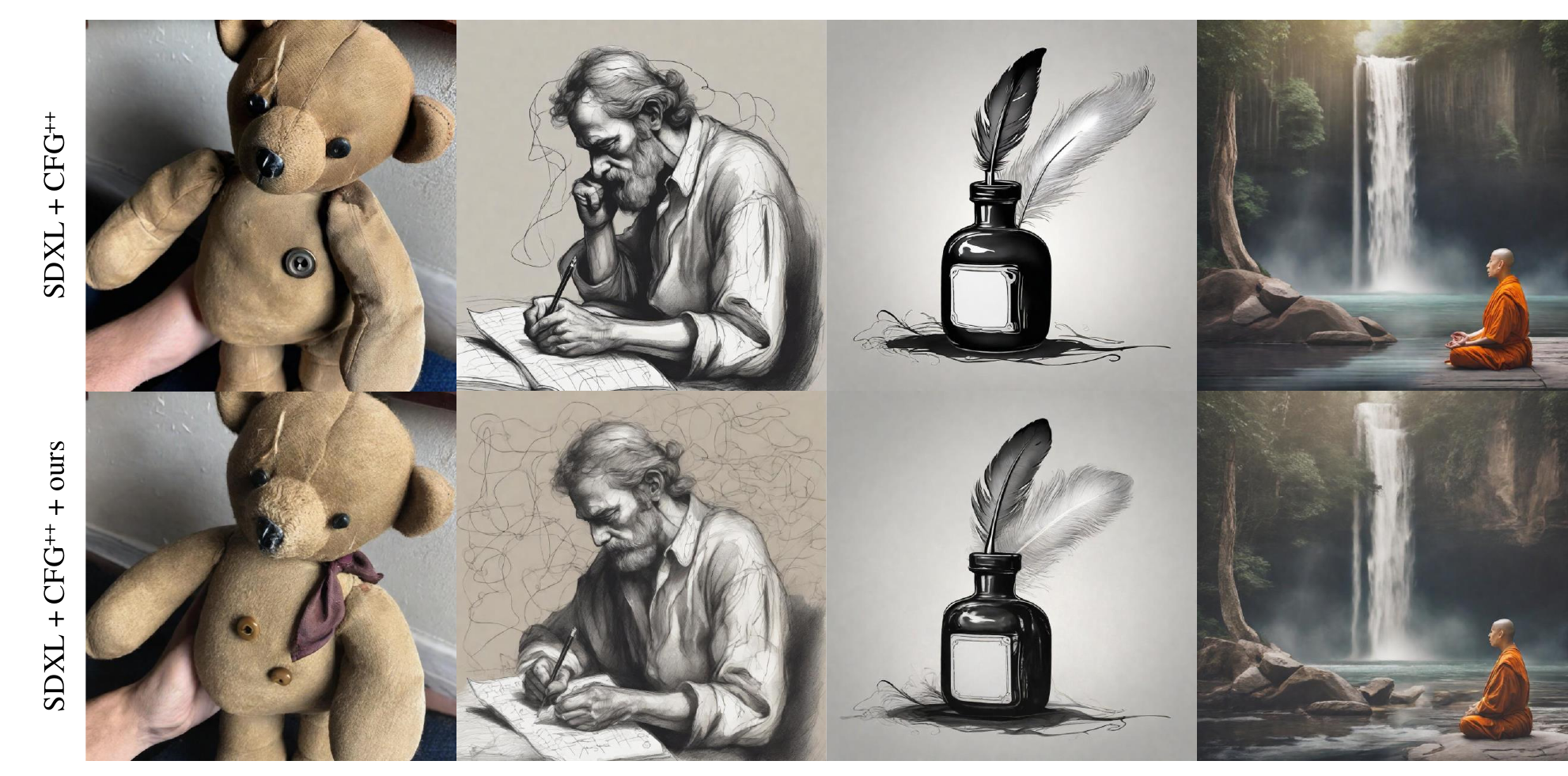}
    \caption{We observed that incorporating our method with CFG++ approaches improved the image structure, details, and overall color quality.}
    \label{fig:cfgpp}
\end{center}
\end{figure*}

CFG++ \cite{chung2024cfg++} proposes an interpolation-based CFG computation method instead of extrapolation. When we applied our projection to the unconditional score used in CFG++, as shown in \fref{fig:cfgpp}, the results improved further. These findings highlight the versatility of our method and its ability to enhance other existing techniques.

\begin{table}[t]
\resizebox{0.8\linewidth}{!}{%
\begin{tabular}{lcc}
                                                         & FID            & CLIPScore     \\ \hline
\multicolumn{1}{l|}{SDXL turbo}                          & 21.47          & 0.31          \\
\multicolumn{1}{l|}{SDXL turbo + ours}                   & \textbf{20.36} & \textbf{0.32} \\ \hline
\multicolumn{1}{l|}{InstaFlow}                           & 16.76          & \textbf{0.30} \\
\multicolumn{1}{l|}{InstaFlow + ours}                    & \textbf{16.19} & \textbf{0.30} \\ \hline
\multicolumn{1}{l|}{PixArt-$\Sigma$}        & 22.53          & \textbf{0.32} \\
\multicolumn{1}{l|}{PixArt-$\Sigma$ + ours} & \textbf{20.19} & \textbf{0.32}
\end{tabular}%
}
\caption{Performance comparison of our method applied to SDXL Turbo, InstaFlow, and PixArt-$\Sigma$. FID scores decrease while CLIPScore remains the same or improves, confirming the broad applicability of our method across different generation models, including high-resolution models.}
\label{tab:additional}
\end{table}

\section{Experimental details.}
We provide details on the sampler, guidance scale, sampling steps, and additional existing baselines' hyperparameters in \tref{tab:details}.

\begin{table}[t]
\resizebox{\linewidth}{!}{%
\begin{tabular}{c|cccc}
Model      & Scheduler                       & CFG scale & Sampling steps & etc              \\ \hline
SD v1.4    & PNDMScheduler                   & 7.5       & 50             & SAG scale: 0.75  \\
SD v1.5    & PNDMScheduler                   & 7.5       & 50             & PAG scale: 3.0   \\
SDXL       & EulerDiscreteScheduler          & 5.0       & 50             & CFG++ scale: 0.6 \\
SD v3      & FlowMatchEulerDiscreteScheduler & 7.0       & 28             &                  \\
SDXL Turbo & EulerAncestralDiscreteScheduler & 2.0       & 1              &                  \\
InstaFlow  & PNDMScheduler                   & 7.5       & 1              &                  \\
PixArt-$\Sigma$   & DPMSolverMultistepScheduler     & 4.5       & 20             &                 
\end{tabular}%
}
\vspace{-1em}
\caption{Experimental details.}
\vspace{-1em}
\label{tab:details}
\end{table}

\section{Additional Results: Few-Step and High-Resolution Image Generation}
We further report the application of our method to few-step generation models and high-resolution image generation. \tref{tab:additional} presents the results when our method is applied to SDXL Turbo (a one-step generation model) and InstaFlow (also a one-step generation model). In both cases, FID scores improve, while CLIPScore remains the same or improves, demonstrating that our method performs effectively not only in many-step models but also across all models utilizing CFG. Notably, for SDXL Turbo, the CFG scale was set to a very low value of 1.3.

\begin{figure*}[!t]
\begin{center}
    \centering
    \includegraphics[width=\textwidth]{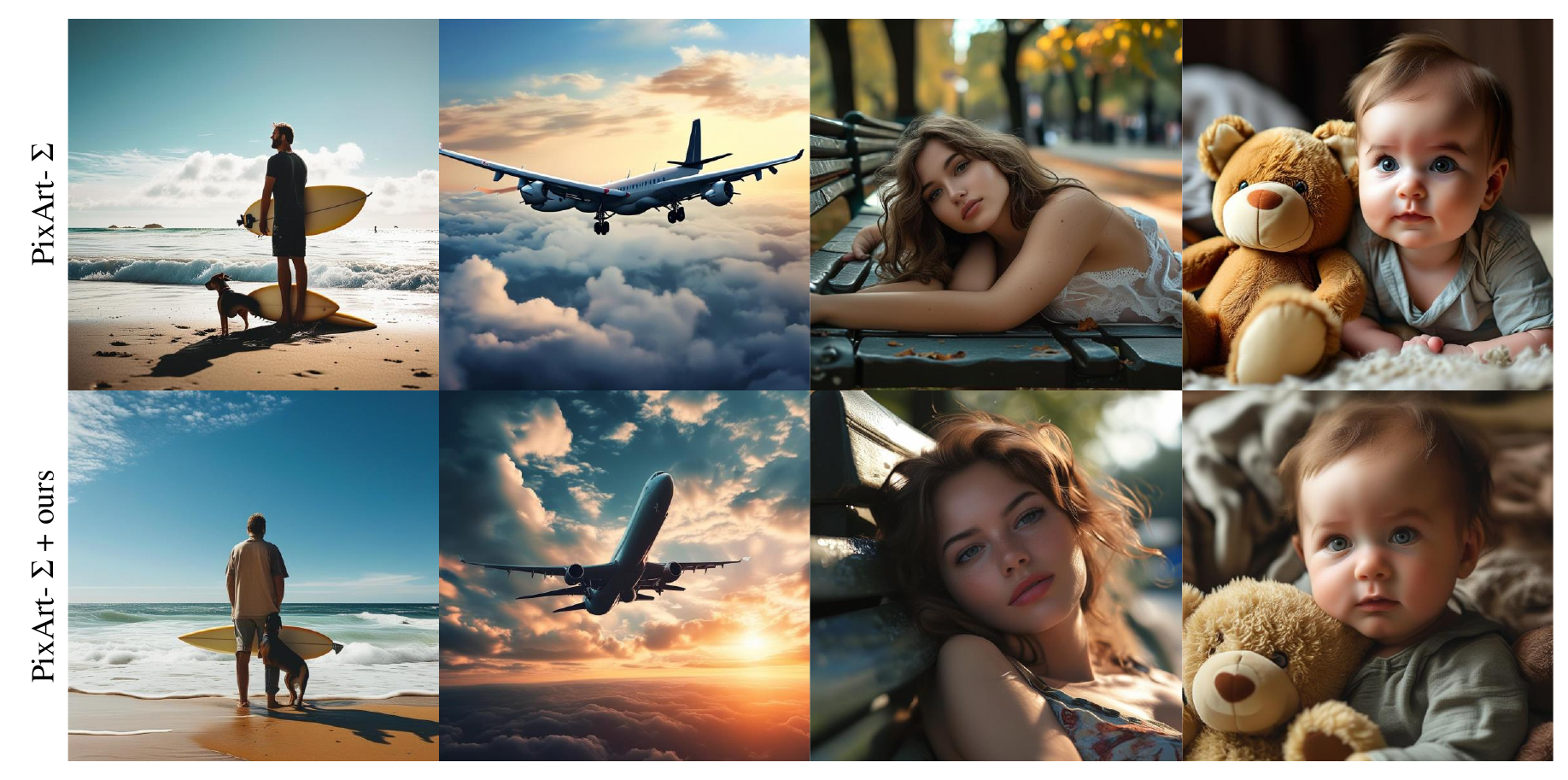}
    \caption{Visual examples generated by PixArt-$\Sigma$ with our method, demonstrating improved image quality in terms of structure, details, and overall aesthetics}
    \label{fig:pixart}
\end{center}
\end{figure*}

Additionally, \tref{tab:additional} highlights the performance of our method in PixArt-$\Sigma$, a high-resolution text-to-image generation model. Similar improvements are observed, with a reduction in FID scores and maintenance of CLIPScore. \fref{fig:pixart} showcases the visual results of PixArt-$\Sigma$, further validating the effectiveness of our approach.

\begin{lstlisting}[language=Python, caption={Code for TCFG with the Hugging Face code style.}, float]
if self.do_classifier_free_guidance:
    noise_pred_uncond, noise_pred_text = noise_pred.chunk(2)

    all_noise = torch.stack((noise_pred_text, noise_pred_uncond), dim=1).to(dtype=torch.float32)
    all_noise = all_noise.reshape(all_noise.size(0), all_noise.size(1), -1)

    U, S, Vh = torch.linalg.svd(all_noise, full_matrices=False)
    Vh = Vh.to(all_noise.device)
    Vh_modified = Vh.clone().to(all_noise.device)
    Vh_modified[:,1] = 0
    noise_null_flat = noise_pred_uncond.reshape(noise_pred_uncond.size(0), 1, -1).to(dtype=torch.float32)
    noise_null_flat = noise_null_flat.to(Vh.device)
    x_Vh = torch.matmul(noise_null_flat, Vh.transpose(-2, -1))
    x_Vh_V = torch.matmul(x_Vh, Vh_modified)
    noise_pred_uncond = x_Vh_V.reshape(*noise_pred_uncond.shape).to(noise_pred_text.dtype).to(noise_pred_text.device)
    noise_pred = noise_pred_uncond + self.guidance_scale * (noise_pred_text - noise_pred_uncond)
\end{lstlisting}

\end{document}